%% file: cvpr_2022.tex
\documentclass[10pt,twocolumn,letterpaper]{article}

\usepackage{cvpr}              

\usepackage{graphicx}
\usepackage{amsmath}
\usepackage{amssymb}
\usepackage{booktabs}

\usepackage{bbold}
\usepackage{float}
\usepackage{bbm}
\usepackage{subcaption}
\usepackage{upgreek}
\usepackage{comment}
\usepackage{bm}
\usepackage[table]{xcolor}

\usepackage[pagebackref,breaklinks,colorlinks]{hyperref}

\usepackage[capitalize]{cleveref}
\crefname{section}{Sec.}{Secs.}
\Crefname{section}{Section}{Sections}
\Crefname{table}{Table}{Tables}
\crefname{table}{Tab.}{Tabs.}

\def\cvprPaperID{2327} 
\def\confName{CVPR}
\def\confYear{2022}

\begin{document}

\title{What Stops Learning-based 3D Registration from Working in the Real World?}

\author{Zheng Dang\\
CVLab, EPFL\\
Switzerland\\
{\tt\small zheng.dang@epfl.ch}
\and
Lizhou Wang\\
Xi'an Jiaotong University\\
China\\
{\tt\small dzyxwanglizhou@stu.xjtu.edu.cn}
\and
Minglei Lu\\
Xi'an Jiaotong University\\
China\\
{\tt\small am12345@stu.xjtu.edu.cn }
\and
Junning Qu\\
Xi'an Jiaotong University\\
China\\
{\tt\small junning.qiu@stu.xjtu.edu.cn }
\and
Mathieu Salzmann\\
CVLab, EPFL \& Clearspace\\
Switzerland\\
{\tt\small mathieu.salzmann@epfl.ch}
}
\maketitle

\input{tex/def}
\input{tex/0-abstract}
\input{tex/1-intro}
\input{tex/2-related}

\input{tex/3-secrets}
\input{tex/4-experiment}
\input{tex/5-conclusion}

{\small
\bibliographystyle{ieee_fullname}
\bibliography{bibtex/string,bibtex/vision}
}

\end{document}

%% file: tex/def.tex
\definecolor{mapillarygreen}{RGB}{38,235,179}

\definecolor{blue1}{RGB}{86,174,139}

\newcommand{\MS}[1]{\textcolor{red}{{\bf #1}}}
\newcommand{\ms}[1]{\textcolor{red}{#1}}
\newcommand{\ZD}[1]{{\color{blue}{\bf ZD: #1}}}
\newcommand{\zd}[1]{{\color{blue}{#1}}}
\newcommand{\YG}[1]{{\color{orange}{\bf JNQ: #1}}}
\newcommand{\yg}[1]{{\color{orange}{#1}}}
\newcommand{\LML}[1]{{\color{green}{\bf LML: #1}}}
\newcommand{\lml}[1]{{\color{green}{#1}}}
\newcommand{\LZW}[1]{{\color{cyan}{\bf LZW: #1}}}
\newcommand{\lzw}[1]{{\color{cyan}{#1}}}

\newcommand{\bX}{\mathcal{X}}
\newcommand{\btX}{\tilde{\mathcal{X}}}
\newcommand{\bx}{\mathbf{x}}
\newcommand{\bbx}{\bar{\mathbf{x}}}
\newcommand{\bY}{\mathcal{Y}}
\newcommand{\by}{\mathbf{y}}
\newcommand{\bby}{\bar{\mathbf{y}}}
\newcommand{\bfx}{\mathbf{f}^x}
\newcommand{\bfy}{\mathbf{f}^y}
\newcommand{\bthx}{\theta^x}
\newcommand{\bthy}{\theta^y}
\newcommand{\bK}{\mathbf{K}}
\newcommand{\bT}{\mathcal{T}}
\newcommand{\bt}{\mathbf{t}}
\newcommand{\bS}{\mathcal{S}}
\newcommand{\bbS}{\mathcal{\bar{S}}}
\newcommand{\bP}{\mathcal{P}}
\newcommand{\bbP}{\mathcal{\bar{P}}}
\newcommand{\bM}{\mathcal{M}}
\newcommand{\bbM}{\mathcal{\bar{M}}}
\newcommand{\bA}{\mathcal{A}}
\newcommand{\bH}{\mathbf{H}}
\newcommand{\bR}{\mathbf{R}}
\newcommand{\bW}{\mathbf{W}}
\newcommand{\bU}{\mathbf{U}}
\newcommand{\bI}{\mathbf{I}}
\newcommand{\bV}{\mathbf{V}}
\newcommand{\ba}{\mathbf{a}}
\newcommand{\bb}{\mathbf{b}}
\newcommand{\bg}{\mathbf{g}}

\newcommand{\hR}{\hat{R}}
\newcommand{\hht}{\hat{\textbf{t}}}
\newcommand{\gR}{R_{gt}}
\newcommand{\gt}{\textbf{t}_{gt}}
\newcommand{\norm}[1]{\left\lVert#1\right\rVert}
\newcommand{\argmin}{\mathop{\mathrm{argmin}}}

\newcommand\largeheight{0.09\columnwidth}

%% file: tex/0-abstract.tex
\begin{abstract}
Much progress has been made on the task of learning-based 3D point cloud registration, with existing methods yielding outstanding results on standard benchmarks, such as ModelNet40, even in the partial-to-partial matching scenario. Unfortunately, these methods still struggle in the presence of real data. 
In this work, we identify the sources of these failures, analyze the reasons behind them, and propose solutions to tackle them. We summarise our findings into a set of guidelines and demonstrate their effectiveness by applying them to different baseline methods, DCP and IDAM. In short, our guidelines improve both their training convergence and testing accuracy.
Ultimately, this translates to a best-practice 3D registration network (BPNet), constituting the first learning-based method able to handle previously-unseen objects in real-world data. Despite being trained only on synthetic data, our model generalizes to real data without any fine-tuning, reaching an accuracy of up to $67\%$ on point clouds of unseen objects obtained with a commercial sensor.

\vspace{-1em}
\end{abstract}

%% file: tex/1-intro.tex
\section{Introduction}
3D registration aims to determine the rigid transformation, i.e., 3D rotation and 3D translation, between two 3D point sets. Over the years, the researchers in this field have actively designed numerous strategies to address this task, the most recent ones focusing on developing deep learning frameworks~\cite{Aoki19,Wang19e,Wang19f,Yew20,Yuan20,Li20}.
While these learning-based approaches yield almost perfect results on synthetic datasets, such as ModelNet40~\cite{Wu15}, their performance on real-world data, such as TUD-L~\cite{Hodan18} or HomeBrew~\cite{Kaskman19} is virtually never reported.

This work was motivated by our unsuccessful attempts at exploiting the state-of-the-art deep learning frameworks for 3D registration on depth data from real scenes. In this context, we observed these frameworks to struggle with the following challenges. First, in real-world data, the sensor may observe the target object from any viewpoint, covering essentially the full rotation range. This significantly contrasts with the $45^\circ$ rotation range typically used in synthetic datasets. 
Second, real-world data contains noise that significantly differs from the Gaussian noise commonly added to synthetic data to simulate sensor noise. Specifically, the real-world noise comes not only from the sensor noise itself but also from the fact that the captured 3D points do not exactly correspond to the model ones because the sensor's sampling pattern is discrete.

In this paper, we therefore seek to understand what stops learning-based 3D registration methods from working in the presence of such real-world challenges. To this end, we conduct a comprehensive and systematic study on the loss functions, normalization techniques, and outlier rejection strategies used in different deep learning frameworks. This allows us to identify the key weaknesses of these frameworks and propose solutions to overcome them. As a result, we propose a best-practice 3D registration method based on the principles derived from our analysis and demonstrate its effectiveness on several real-world datasets. 

Our contributions can be summarized as follows:
\begin{itemize}
    \setlength\itemsep{.05em}
    \item We identify the issues that prevent existing learning-based methods from working on real-world data, analyze the reasons behind them and propose solutions to address them.
    \item We summarize our findings in a few guidelines to help others modify their own method and make it effective in the presence of real data. 
    \item We evidence the effectiveness of our guidelines by applying them to two baseline methods, DCP~\cite{Wang19e} and IDAM~\cite{Li20}. In both cases, we observe a significant performance improvement on real-world datasets after modification.
    \item Ultimately, our approach constitutes the first learning-based method able to handle previously-unseen objects in real-world data. Despite being trained on synthetic data only, this model generalizes to real data \textbf{without} any fine-tuning, reaching an accuracy of up to $67\%$ on point clouds of new objects acquired with a commercial sensor.
\end{itemize}
We demonstrate the generality of our analysis and of our solutions by performing experiments on three real datasets: our own high-quality XEPose dataset, TUD-L~\cite{Hodan18} and HomeBrew~\cite{Kaskman19}. Our code will be made publicly available.

%% file: tex/2-related.tex
\section{Related Work}

\textbf{Traditional point cloud registration.} Although we focus on learning-based 3D registration, many non-learning algorithms have been proposed in the past, Iterative Closest Point (ICP) being the best-known one. 
Several variants, such as Generalized-ICP~\cite{Segal09} and Sparse ICP~\cite{Bouaziz13}, have been proposed to improve robustness to noise and mismatches, and we refer the reader to~\cite{Pomerleau15,Rusinkiewicz01} for a complete review of ICP-based strategies. The main drawback of these methods is their requirement for a reasonable initialization to converge to a good solution. Only relatively recently has this weakness been addressed by the globally-optimal registration method Go-ICP~\cite{Yang15}. In essence, this approach follows a branch-and-bound strategy to search the entire 3D motion space $SE(3)$. 
While globally optimal, Go-ICP comes at a much higher computational cost than vanilla ICP. This was, to some degree, addressed by the Fast Global Registration (FGR) algorithm~\cite{Zhou16}, which leverages a local refinement strategy to speed up computation. While effective, FGR still suffers from the presence of noise and outliers in the point sets, particularly because, as vanilla ICP, it relies on 3D point-to-point distance to establish correspondences. In principle, this can be addressed by designing point descriptors that can be more robustly matched. For example, \cite{Vidal18} relies on generating pose hypotheses via feature matching, followed by a RANSAC-inspired method to choose the candidate pose with the largest number of support matches. Similarly, TEASER~\cite{Yang19} and its improved version TEASER++~\cite{Yang20} take putative correspondences obtained via feature matching as input and remove the outlier ones by an adaptive voting scheme. Over the years, several works have tackled the point matching task for 3D registration,
in both a non learning-based~\cite{Johnson99,Rusu08,Rusu09} and learning-based~\cite{Zeng17b,Khoury17} fashion.
Nowadays, however, these approaches are typically outperformed by end-to-end learning frameworks, which directly take the point sets as input.


\textbf{End-to-end learning with point sets.} 
A key requirement to enable end-to-end learning-based registration was the design of deep networks acting on unstructured sets. Deep sets~\cite{Zaheer17} and PointNet~\cite{Qi17} constitute the pioneering works in this direction. They use shared multilayer perceptrons to extract high-dimensional features from the input point coordinates, and exploit a symmetric function to aggregate these features. This idea was then extended in PointNet++~\cite{Qi17a} via a modified sampling strategy to robustify the network to point clouds of varying density, and in DGCNN~\cite{Wang18b} by building a graph over the point cloud. While the above-mentioned works focus on other tasks than us, such as point cloud classification or segmentation, end-to-end learning for registration has recently attracted a growing attention. In particular, PointNetLK~\cite{Aoki19} combines the PointNet backbone with the traditional, iterative Lucas-Kanade (LK) algorithm~\cite{Lucas81} so as to form an end-to-end registration network; DCP~\cite{Wang19e} exploits DGCNN backbones followed by Transformers~\cite{Vaswani17} to establish 3D-3D correspondences, which are then passed through an SVD layer to obtain the final rigid transformation. While effective, PointNetLK and DCP cannot tackle the partial-to-partial registration scenario. That is, they assume that both point sets are fully observed, during both training and test time. This was addressed by PRNet~\cite{Wang19f} via a deep network designed to extract keypoints from each input set and match these keypoints. This network is then applied in an iterative manner, so as to increasingly refine the resulting transformation. IDAM~\cite{Li20} builds on the same idea as PRNet, using a two-stage pipeline and a hybrid point elimination strategy to select keypoints.
By contrast, RPM-Net~\cite{Yew20} builds on DCP and adopts a different strategy, replacing the softmax layer with an optimal transport one so as to handle outliers. Nevertheless, as PRNet, RPM-Net relies on an iterative strategy to refine the computed transformation. DeepGMR~\cite{Yuan20} leverages mixtures of Gaussians and formulates registration as the minimization of the KL-divergence between two probability distributions to handle outliers.
In any event, the methods discussed above were designed to handle point-clouds in full 3D, and were thus neither demonstrated for registration from 2.5D measurements, nor evaluated on real scene datasets, such as TUD-L and HomebrewedDB. In this work, we identify and solve the issues that prevent the existing learning-based methods from working on real-world data, and, as a result, develop the first learning-based point cloud registration method able to handle  previously-unseen objects in real data captured with a commercial sensor.

%% file: tex/3-secrets.tex
\section{Analysis of Learning-based 3D Registration}
\input{fig/overview}

In this section, we analyze different aspects of learning-based 3D registration frameworks. For our analysis, we focus on DCP(v2)~\cite{Wang19e} as a baseline framework because of its simplicity and because it has served as starting point for several works~\cite{Wang19f,Yew20,Li20}. As will be evidenced by our experiments, however, our conclusions generalize to other frameworks. Below, we first introduce our baseline framework and experimental settings, and then turn to our analysis.


\subsection{Analysis Setup}
\label{baseline}
\paragraph{Baseline Method.} Let us first briefly review the DCP(v2)~\cite{Wang19e} pipeline, which serves as our baseline for our analysis. To this end, let $\bX \in \mathbb{R}^{M \times 3}$ and $\bY \in \mathbb{R}^{N \times 3}$ be two sets of 3D points sampled from the same object surface. We typically refer to $\bX$ as the source point set and to $\bY$ as the target point set. Registration then aims to find a rigid transformation $\bT$ that aligns $\bX$ to $\bY$. DCP(v2) combines a DGCNN~\cite{Wang18b} with a Transformer~\cite{Vaswani17}. Specifically, the DGCNN takes a point set as input and outputs point-wise features. Let $\bthx$, resp. $\bthy$, be the final feature matrix, i.e., one $P$-dimensional feature vector per 3D point, for $\bX$, resp. $\bY$. The transformer takes the features $\bthx$ and $\bthy$ as input and learns a function $\phi : \mathbb{R}^{M\times P} \times \mathbb{R}^{N\times P} \rightarrow \mathbb{R}^{M\times P}$, which combines the information of the two point sets. Ultimately, this produces a descriptor matrix $\bfx$, resp. $\bfy$, for $\bX$, resp. $\bY$. Given these matrices, DCP(v2) then forms a score map $\bS \in \mathbb{R}^{M\times N}$ by computing the similarity between each source-target pair of descriptors. That is, we compute the $(i,j)$-th element of $\bS$ as
\begin{equation}
    \bS_{i,j} = <\bfx_i, \bfy_j>, \;\;\forall (i, j) \in [1,M]\times [1,N]\;,
\end{equation}
where $<\cdot, \cdot>$ is the inner product, and $\bfx_i, \bfy_j \in \mathbb{R}^P$. 

This score map is passed through a row-wise SoftMax so as to obtain correspondences. These correspondences are then processed within the network via an SVD layer to solve the Procrustes problem, and the resulting rigid transformation is compared to the ground-truth one with a mean squared error (MSE) loss.
\vspace{-1.em}

\paragraph{Datasets.} 
Although our goal utlimately is to design an effective method for real data, we perform parts of our analysis on synthetic one, which lets us move step by step towards the real scenario.  To this end, we follow the PRNet~\cite{Wang19f} experimental setup for partial-to-partial registration using the auto-aligned ModelNet40 dataset~\cite{Wu15,Sedaghat16}. 
This dataset~\cite{Wu15} contains 12,311 CAD models spanning 40 object categories. As in~\cite{Wang19f}, we split it into 9,843 training and 2,468 testing samples. The point clouds in this dataset are normalized in the range $[-1, 1]$ on each axis. During training, we sample 1024 points from the original point clouds to define the source point set $\bX$. 
We then sample a random rigid transformation along each axis, with rotation in $[0^{\circ}, 45^{\circ}]$ and translation in $[-0.5, 0.5]$, and apply it to the point set $\bX$ to obtain the target point set $\bY$. Furthermore, we also evaluate the full rotation range scenario, where we set the rotation range (in Euler andgles) to $[-180^{\circ}, 180^{\circ}]$ for the $x$, and $z$ axes, and to $[-90^{\circ}, 90^{\circ}]$ for the $y$ axis.
To simulate partial scans of $\bX$ and $\bY$, as in PRNet~\cite{Wang19f}, we randomly select one point in $\bX$ and one point in $\bY$ independently, and only keep their 768 nearest neighbors.

In addition to this synthetic data, we analyse the behavior of 3D registration on real data, with self-occluded and potentially unseen objects. To this end, and to discard the effects due to noisy measurements, we acquired our own dataset, which we dubbed XEPose, with a high-precision depth sensor. The XEPose dataset contains 2 sequences depicting 5 objects simultaneously captured with a precision of up to 20$um$.
We used the ground-truth mask of each object to extract their depth measurements, from which we formed a target point cloud by using the provided camera parameters. 
For all experiments on real data, we discarded all objects having fewer than 4096 points and a visibility ratio smaller than 90 percent. This left us with 6428 object instances for the XEPose dataset.

\input{table/batch_norm}
\input{table/loss_func}

\vspace{-1.em}
\paragraph{Evaluation Protocols.}
\label{eval_protocols}

We perform experiments on both synthetic data (ModelNet40) and real-world data (XEPose, TUD-L, HomebrewedDB), and thus use two different training and evaluation protocols for these two scenarios. In the case of synthetic data, 
we evaluate the models on the partial-to-partial point cloud registration task. Specifically, we follow the unseen-object setting used in~\cite{Wang19f,Yew20,Li20}. We then report the rotation and translation error between the predictions $\hR,\hht$ and the ground truth $\gR,\gt$. These errors are computed as
\begin{equation}
    \begin{aligned}
        &E_{rot}(\hR,\gR) = arccos\frac{trace(\hR^{\top}\gR) - 1}{2} \;,\\
        &E_{trans}(\hht,\gt) = \frac{1}{n}\sum^{n}_{i=1}(\hat{t}_i - t_i)^{2}\;.
    \end{aligned}
    \label{syn_metric}
\end{equation}
We summarize the results in terms of mean average precision (mAP) of the estimated relative pose under varying accuracy thresholds, as in~\cite{Yi18}. For rotation, we use the thresholds $[5^{\circ}, 10^{\circ}, 15^{\circ}, 20^{\circ}, 25^{\circ}, 30^{\circ}]$. For translation, we set the thresholds to be $[1\times 10^{-3}, 5\times 10^{-3}, 1\times 10^{-2}, 5\times 10^{-2},1\times 10^{-1}, 5\times 10^{-1}]$ in the normalized data.

The real data differs from the synthetic one in that it corresponds to depth maps, i.e., 2.5D measurements instead of 3D. To handle this,
we use an on-the-fly rendering strategy to generate data, as discussed in more detail in Section~\ref{sec:xepose}. Unlike the synthetic data, the models and depth maps of the real datasets correspond to actual physical quantities. Since the translation mean squared error (MSE) in Eq.~\ref{syn_metric} only assesses numerical errors without physical meaning, we replace it with $E_{trans}(\hht,\gt) = \norm{\hht - \gt}^{2}_2$, and set the thresholds to be $[0.5cm, 1cm, 2cm, 5cm, 10cm, 15cm]$. We keep the rotation error unchanged. Furthermore, we also report the average distance metric (ADD)~\cite{Xiang18}, which measures the average distance between the 3D model points transformed using the predicted pose and those obtained with the ground-truth one. We set the threshold to be $10\%$ of the model diameter, as commonly done in 6D pose estimation.

\subsection{Analysis}
\label{secrets}
We now analyze different aspects of learning-based 3D registration. Our analysis covers both data processing techniques, and network design and training strategies.

\input{table/matches_selection_layer}
\input{table/voxelization}
\paragraph{Data Processing.}
The first aspect we investigate is the importance of pre-processing the data. We study two data processing steps, which may lead to poor performance if not performed, and provide corresponding recommendations. 

Let us first look into the scale normalization of the object's mesh model. The scale of the model depends on both the unit (e.g., meter vs millimeter) used to describe the object and the actual object size (e.g., a car is larger than a cup). As evidenced by the ModelNet40 results w/o scale normalization 
in Table~\ref{tab:batch_norm}, failing to normalize the object leads to poor accuracy because of the too large variations in coordinates that may occur in the dataset. Note that, because the objects in ModelNet40 are already normalized, this problem has never been highlighted in the past. However, real objects typically aren't normalized, and we therefore recommend normalizing their coordinates into a cube in the range $[-1, 1]$. Unless otherwise specified, all remaining experiments will use this normalization. After prediction, we transfer the results back to the original scale to obtain the final pose. 

The second processing step focuses on the translation of the target point cloud. In the synthetic data scenario, this never comes as an issue, as the target point cloud is generated within a small translation range of the source point cloud. However, this is not the case with real data, and processing the raw data using a deep network suffers from the fact that large variations can be observed across the different object instances, particularly for the translation in the $z$ (i.e., depth) direction. For methods that, as our baseline, directly process the target 3D coordinates, we therefore recommend subtracting the mean of the 3D coordinates from the target point cloud. After removing the translation, we then scale the target point cloud using the same scale values as for the corresponding mesh model. 
\vspace{-1.25em}
\paragraph{Batch Normalization Strategy.}
The second aspect we study relates to the batch normalization operations performed in the network. 
The standard batch normalization process relies on keeping track of running means and variances of the features throughout the training process, and then use these statistics for evaluation. While this works well when the shapes of the objects do not differ much, ModelNet40 contains 9,843 training objects, which differ significantly in shape. In this scenario, the statistics of the data in current batch can depart drastically from those of the last batch. We observed this to slow down the convergence of the network. Furthermore, the difference between the statistics recorded in the training process and those of the current evaluation batch also lead to a decrease pose estimation accuracy. 

To overcome this, we propose to use the current batch statistics during both training and evaluation. 
The benefits of this strategy are shown in Table~\ref{tab:batch_norm}, where we compare the number of training iterations as well as the rotation and translation mAP obtained using the running statistics and the current statistics in batch normalization. The use of current statistics outperforms that of running ones for all rotation thresholds.
Furthermore, it makes the network converge in only $10,000$ iterations, while using the running statistics requires $60,000$ iterations at least.

\input{fig/xepose_qual}

\paragraph{Loss Function.}
\label{sec:nll_loss}
In the commonly-used synthetic setting, the relative rotation between the two point clouds is limited to the $[0^{\circ}, 45^{\circ}]$ range. By contrast, in real data, the objects' pose may cover the full rotation range. To mimic this, we simply modified the data augmentation setting on ModelNet40 so as to generate samples in the full rotation range. As shown in Table~\ref{tab:loss_func}, we then observed our baseline to fail in this setting, even when using the data processing steps and batch normalization strategy we recommended above.

Via a detailed analysis of the training behavior, we traced the reason for this failure back to the choice of loss function. Specifically, the use of an SVD-based loss function yields instabilities in the gradient computation. As can be seen from the mathematical expression of the SVD derivatives in~\cite{Ionescu15}, when two singular values are close to each other in magnitude, the derivatives explode, thus interrupting the training process. 
To cope with this problem, inspired by~\cite{Yi18}, we propose to use the negative log likelihood loss to impose a direct supervision on the score map.
As shown in Table~\ref{tab:loss_func}, the NLL loss outperforms the SVD+MSE one by a large margin in the $45^\circ$ rotation range, and remains effective in the full rotation-range scenario, where the SVD+MSE combination fails.

\vspace{-1.3em}
\paragraph{Outlier Rejection Strategy.}
\label{sec:outlier_rejection}
The original DCP architecture uses the Softmax operator on the score map to establish correspondences. This implicitly assumes that each 3D point in the target point set can always be matched to a source (model) point. However, this assumption is violated in the presence of outliers in the target point set. To account for outliers, RPMNet~\cite{Yew20} employs a Sinkhorn layer with an outlier bin. However, in our initial experiments, we observed the Sinkhorn layer to yield worse results than the Softmax operation.

We found the reason behind to be twofold.
First, with synthetic data, the point clouds overlap heavily, between $50\%$ to $100\%$, which is well handled by the Softmax operation.
Second, RPMNet relies on a weighted SVD strategy that cannot leverage the extended score map output by the Sinkhorn layer as a weight matrix to calculate the final pose. In other words, the outlier bin is essentially dropped from the loss computation and thus not updated by back-propagation. 


To address these issues, we exploit a hard selection strategy and use the NLL loss to supervise the score map including the outlier bin. Specifically, we use the output of the Sinkhorn layer to find the best set of corresponding points between two point clouds. In addition to the points found as outliers, we also purge those whose value in the score map are below a threshold. As shown in Table~\ref{tab:matches_layer}, this strategy outperforms the weighted SVD one by a large margin. We will show later that our strategy also outperforms the keypoint selection approach used in PRNet~\cite{Wang19f} and IDAM~\cite{Li20}.

\input{table/xepose}

\vspace{-1.25em}
\paragraph{Correspondence Bijection.}
\label{sec:data_norm}

In an ideal scenario, when the outliers have been removed, the source and target point clouds should be in bijection. However, we observed this never to be satisfied when working with real data. An important reason behind this is the fact that the point clouds are inhomogeneously distributed; their distribution depends on the sensor's sampling pattern. For example, an object with two surfaces of same size and shape but viewed by the sensor from a different angle will not have the same point density on both surfaces. Such an inhomogeneity may cause one point not to have a perfect correspondence in the other point set, but rather to roughly match multiple points within a small area. The Sinkhorn layer will then choose one such imperfect correspondence, which may deteriorate the pose estimation result.

To avoid this, we perform a voxel-based downsampling of the point clouds. Specifically, we downsample both the source and target point clouds using the same voxel size to ensure that the density between the two point clouds is similar.
Furthermore, to reflect the fact that the target point cloud is only a partial observation of the mesh model, we set the number of points in the target point cloud to be less than that of the source point cloud. 
As shown in Table~\ref{tab:data_norm}, where we compare different versions of our recommended strategies on the real XEPose dataset, voxel downsampling yields a further improvement. We will further evidence this on other, more noisy real datasets in Section~\ref{sec:experiments}.
\vspace{-1.em}
\paragraph{Summary.}
Altogether, our analysis has evidenced the importance of
\vspace{-.5em}
\begin{itemize}
    \setlength\itemsep{.02em}
    \item processing both the model and target point cloud to facilitate network training;
    \item relying on the current batch statistics in batch normalization to speed up convergence and improve the pose estimation accuracy;
    \item exploiting the NLL loss function to handle the full rotation range;
    \item performing hard selection with a Sinkhorn layer to account for outliers;
    \item voxel downsampling the point clouds to obtain bijective correspondences.
\end{itemize}
\vspace{-.4em}
Our experiments have shown that a network combining all such strategies, which we dub \textbf{BPNet}, for Best Practice Network, can effectively perform registration on previously unseen objects in real-world data, whereas the baseline DCP(v2) network fails completely. In the next section, we will show that our recommendation also apply to other frameworks, and that our approach yield state-of-the-art results on real datasets acquired with commercial sensors, and thus noisier than XEPose.

%% file: fig/overview.tex
\begin{figure*}[!ht]
    \centering
    \includegraphics[width=.80\linewidth]{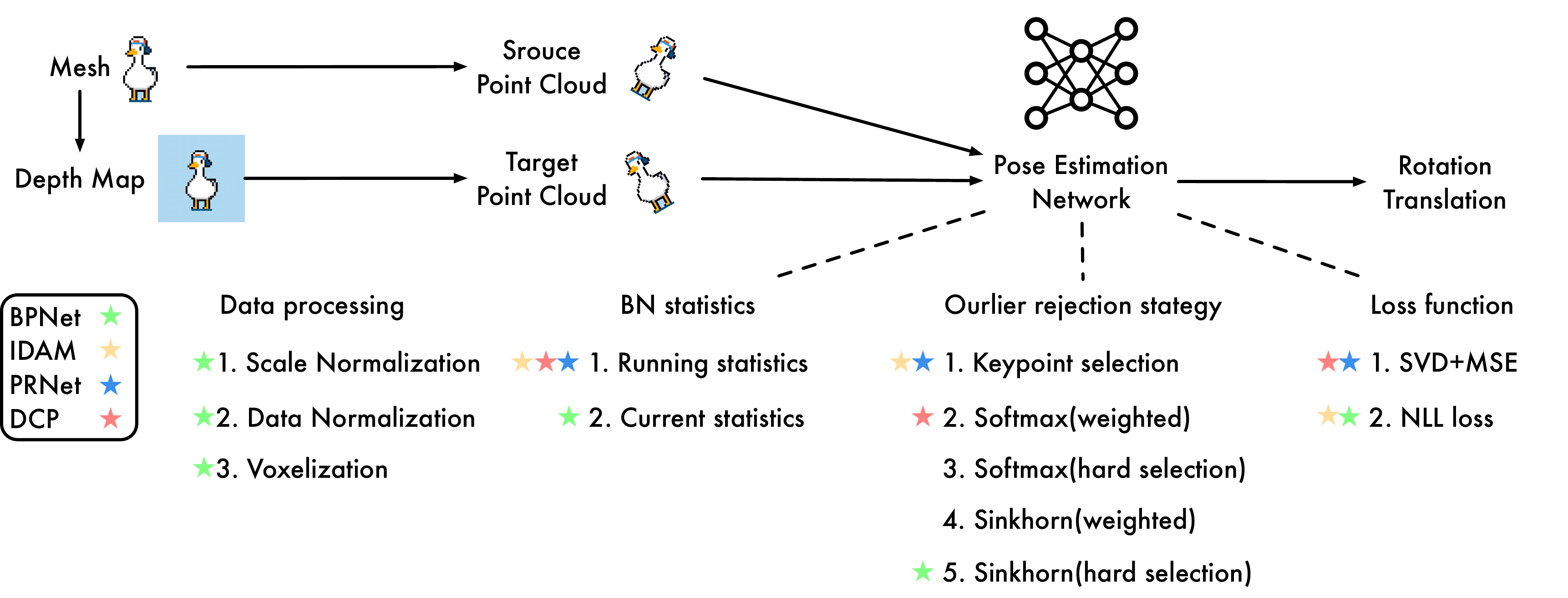}
    \caption{\label{fig:overview} Overview of the aspects we analyze. We show which techniques are used by each method using stars of different colors.}
    \vspace{-1.25em}
\end{figure*}

%% file: table/batch_norm.tex
\begin{table*}[!htb]
    \centering
    {\footnotesize
    \setlength{\tabcolsep}{2.0mm}{
    \begin{tabular}{l|cccccc|cccccc|c}
        \toprule
        & \multicolumn{6}{c|}{Rotation mAP} & \multicolumn{6}{c|}{Translation mAP} & Iters \\
        Method & $5^{\circ}$ & $10^{\circ}$ & $15^{\circ}$ & $20^{\circ}$ & $25^{\circ}$ & $30^{\circ}$ & $0.001$ & $0.005$ & $0.01$ & $0.05$ & $0.1$ & $0.5$ \\
        \midrule
        Vanilla DCP(v2) w/o Scale Normalization & 0.02 & 0.06 & 0.09 & 0.14 & 0.19 & 0.23 & 0.00 & 0.00 & 0.00 & 0.01 & 0.01 & 0.05 & 60000 \\
        Vanilla DCP(v2)  & 0.28 & 0.70 & 0.88 & 0.95 & 0.98 & 0.99 & 0.49 & 0.94 & 0.99 & 1.00 & 1.00 & 1.00 & 60000 \\ 
        DCP(v2)+BN (Current statistics) & 0.36 & 0.75 & 0.90 & 0.96 & 0.98 & 0.99 & 0.51 & 0.92 & 0.98 & 1.00 & 1.00 & 1.00 & 10000 \\ 
        \bottomrule
    \end{tabular}
    }}
    \caption{Influence of data processing and of the batch normalization statistics. ``Vanilla DCP(v2)" uses the running statistics, and ``Vanilla DCP(v2) w/o Scale Normalization'' does not normalize the scale of the mesh models. A comparison of the first two rows indicates the importance of scale normalization. ``DCPv2+BN (Current statistics)'' uses the current batch's statistics. Its much better accuracy than ``Vanilla DCP(v2)" shows the benefits of using the current statistics instead of the running ones for 3D registration. These results were computed on the tasks of partial-to-partial registration on the ModelNet40 dataset and by setting the rotation range to $45^\circ$ on each axis. 
    }
    \label{tab:batch_norm}
\end{table*}

%% file: table/loss_func.tex
\begin{table*}[!htb]
    \centering
    {\footnotesize
    \setlength{\tabcolsep}{2.5mm}{
    \begin{tabular}{l|cccccc|cccccc}
        \toprule
        & \multicolumn{6}{c|}{Rotation mAP} & \multicolumn{6}{c}{Translation mAP} \\
        Method & $5^{\circ}$ & $10^{\circ}$ & $15^{\circ}$ & $20^{\circ}$ & $25^{\circ}$ & $30^{\circ}$& $0.001$ & $0.005$ & $0.01$ & $0.05$ & $0.1$ & $0.5$ \\
        \midrule
        DCP(v2)+BN+SVD ($45^\circ$)   & 0.36 & 0.75 & 0.90 & 0.96 & 0.98 & 0.99 & 0.51 & 0.92 & 0.98 & 1.00 & 1.00 & 1.00 \\ 
        DCP(v2)+BN+NLL ($45^\circ$)   & 0.72 & 0.97 & 1.00 & 1.00 & 1.00 & 1.00 & 0.54 & 0.99 & 1.00 & 1.00 & 1.00 & 1.00 \\ 
        DCP(v2)+BN+SVD (Full) & 0.00 & 0.01 & 0.02 & 0.03 & 0.04 & 0.05 & 0.04 & 0.23 & 0.42 & 0.96 & 1.00 & 1.00 \\ 
        DCP(v2)+BN+NLL (Full) & 0.35 & 0.67 & 0.86 & 0.94 & 0.97 & 0.98 & 0.19 & 0.57 & 0.85 & 1.00 & 1.00 & 1.00 \\ 

        \bottomrule
    \end{tabular}
    }}
    \caption{Influence of the loss function. In this experiment, all models use the current batch statistics and rely on processed data. 
    ``DCP(v2)+BN+SVD($45^\circ$)'' uses the SVD+MSE loss function, and ``DCP(v2)+BN+NLL($45^\circ$)'' the NLL loss function. The models are evaluated on the partial-to-partial registration task on ModelNet40, with either a $45^\circ$ rotation range, or a full one. For a given rotation range, the NLL loss yields better results than the SVD+MSE one. In the full rotation range scenario, the SVD+MSE loss fails completely, while the NLL loss still yields reasonably accurate pose estimates. This evidences the importance of using the NLL loss, particularly to work in the realistic scenario where the object can be observed from any viewpoint.
    }
    \label{tab:loss_func}
    \vspace{-1.25em}
\end{table*}

%% file: table/matches_selection_layer.tex
\begin{table*}[!htb]
    \centering
    {\footnotesize
    \setlength{\tabcolsep}{2.2mm}{
    \begin{tabular}{l|cccccc|cccccc}
        \toprule
        & \multicolumn{6}{c|}{Rotation mAP} & \multicolumn{6}{c}{Translation mAP} \\
        Method & $5^{\circ}$ & $10^{\circ}$ & $15^{\circ}$ & $20^{\circ}$ & $25^{\circ}$ & $30^{\circ}$& $0.001$ & $0.005$ & $0.01$ & $0.05$ & $0.1$ & $0.5$ \\
        \midrule
        DCP(v2)+BN+NLL+Softmax(Weighted)  & 0.26 & 0.56 & 0.80 & 0.91 & 0.96 & 0.97 & 0.18 & 0.64 & 0.88 & 1.00 & 1.00 & 1.00 \\ 
        DCP(v2)+BN+NLL+Softmax(Hard selection)  & 0.35 & 0.67 & 0.86 & 0.94 & 0.97 & 0.98 & 0.19 & 0.57 & 0.85 & 1.00 & 1.00 & 1.00 \\ 
        DCP(v2)+BN+NLL+Keypoint selection  & 0.38 & 0.65 & 0.81 & 0.90 & 0.93 & 0.95 & 0.16 & 0.39 & 0.52 & 0.99 & 1.00 & 1.00 \\ 
        DCP(v2)+BN+NLL+Sinkhorn(Weighted)   & 0.26 & 0.43 & 0.57 & 0.67 & 0.74 & 0.79 & 0.18 & 0.44 & 0.59 & 0.99 & 1.00 & 1.00 \\ 
        DCP(v2)+BN+NLL+Sinkhorn(Hard selection)  & 0.49 & 0.73 & 0.88 & 0.94 & 0.97 & 0.99 & 0.48 & 0.79 & 0.93 & 1.00 & 1.00 & 1.00 \\ 
        \bottomrule
    \end{tabular} 
    }}
    \caption{Influence of the outlier rejection strategy. All models use the current batch statistic and the NLL loss function, and rely on processed data.
    ``DCP(v2)+BN+NLL+Softmax(Weighted)'' uses the Softmax operator with the weighting strategy as in DCP; ``DCP(v2)+BN+NLL+Keypoint selection'' uses the keypoint selection strategy of PRNet; ``DCP(v2)+BN+NLL+Sinkhorn(Weighted)'' uses a Sinkhorn layer with the same weighting strategy as in RPMNet; ``DCP(v2)+BN+NLL+Softmax(Hard selection)'' uses the Softmax operator but with a hard selection strategy. The hard selection removes the low confidence results from the estimates and therefore better copes with noise than the weighted strategy. As a result, even with the same Softmax operator, hard selection outperforms the weighted one. ``DCP(v2)+BN+NLL+Sinkhorn(Hard selection)'' applies the same principle to the Sinkhorn layer, which also yields a significant performance improvement. These results were computed for partial-to-partial registration on ModelNet40 and with the full rotation range. 
    }
    \label{tab:matches_layer}
\end{table*}

%% file: table/voxelization.tex
\begin{table*}[!htb]
    \centering
    {\footnotesize
    \setlength{\tabcolsep}{1.5mm}{
    \begin{tabular}{l|cccccc|cccccc|c}
        \toprule
        & \multicolumn{6}{c|}{Rotation mAP} & \multicolumn{6}{c|}{Translation mAP} & \multicolumn{1}{c}{ADD} \\
        Method & $5^{\circ}$ & $10^{\circ}$ & $15^{\circ}$ & $20^{\circ}$ & $25^{\circ}$ & $30^{\circ}$& $0.5cm$ & $1cm$ & $2cm$ & $5cm$ & $10cm$ & $15cm$ & $0.1$\\
        \midrule
        DCPv2+BN+NLL+Sinkhorn         & 0.00 & 0.01 & 0.04 & 0.06 & 0.09 & 0.11 & 0.01 & 0.07 & 0.25 & 0.81 & 1.00 & 1.00 & 0.04 \\ 
        DCPv2+BN+NLL+Sinkhorn+Norm              & 0.17 & 0.47 & 0.62 & 0.70 & 0.75 & 0.78 & 0.35 & 0.81 & 0.98 & 1.00 & 1.00 & 1.00 & 0.68 \\ 
        DCPv2+BN+NLL+Sinkhorn+Norm+Voxel  & 0.30 & 0.55 & 0.71 & 0.81 & 0.86 & 0.90 & 0.52 & 0.84 & 1.00 & 1.00 & 1.00 & 1.00 & 0.77\\ 

        \bottomrule
    \end{tabular}
    }}
    \caption{Influence of data processing and voxelization. Here, we evaluate the methods on the XEPose real scene dataset. ``DCPv2+BN+NLL+Sinkhorn+Norm'' uses data processing to move the input coordinates to a more reasonable range than ``DCPv2+BN+NLL+Sinkhorn'', which does not give satisfactory results on this real data.
    The use of voxel downsampling enforces better correspondence bijection, leading to a further boost in the results.
    }
    \label{tab:data_norm}
    \vspace{-1.25em}
\end{table*}

%% file: fig/xepose_qual.tex
\begin{figure*}[!htb]
    \centering
    \begin{subfigure}{.16\textwidth}{\centering\includegraphics[width=\linewidth]{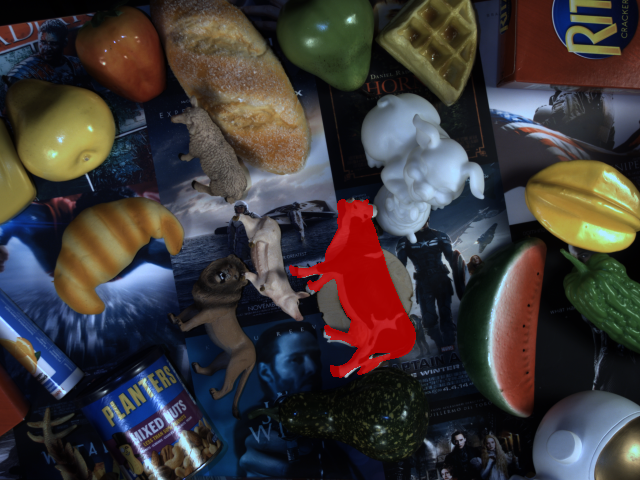}}\end{subfigure}
    \begin{subfigure}{.16\textwidth}{\centering\includegraphics[width=\linewidth]{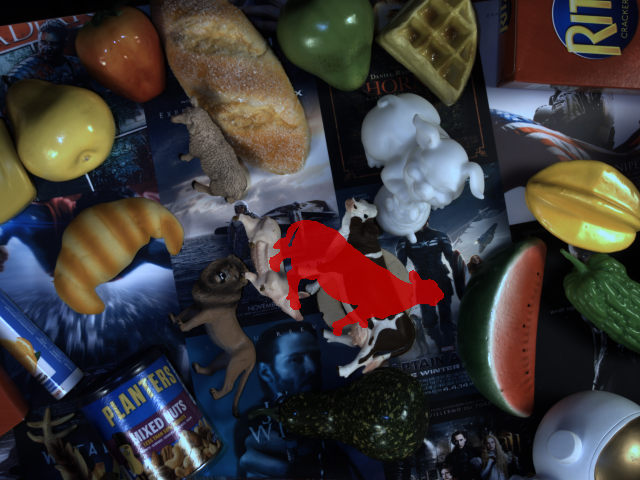}}\end{subfigure}
    \begin{subfigure}{.16\textwidth}{\centering\includegraphics[width=\linewidth]{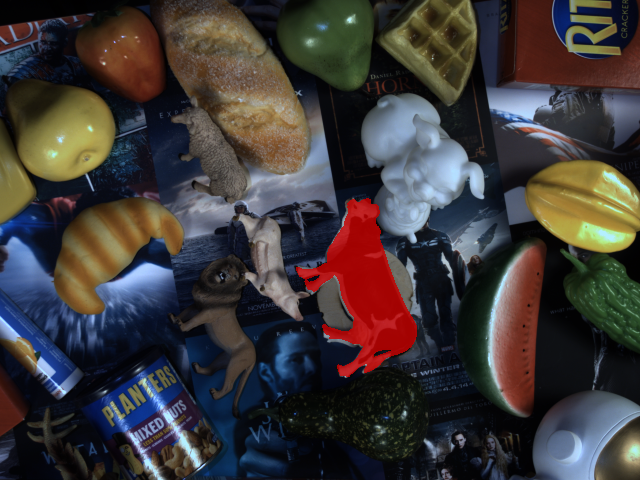}}\end{subfigure}
    \begin{subfigure}{.16\textwidth}{\centering\includegraphics[width=\linewidth]{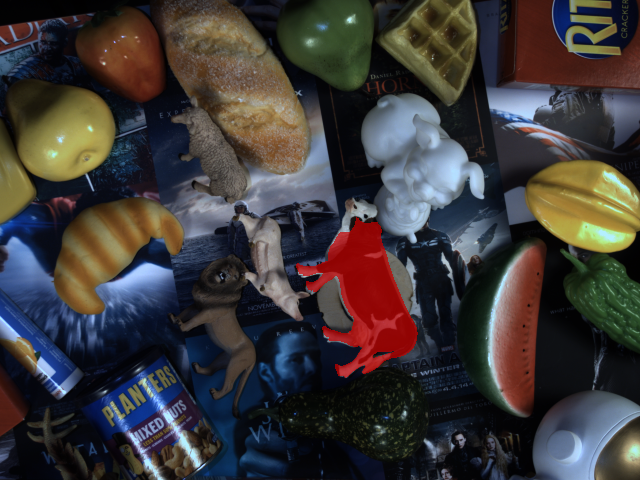}}\end{subfigure}
    \begin{subfigure}{.16\textwidth}{\centering\includegraphics[width=\linewidth]{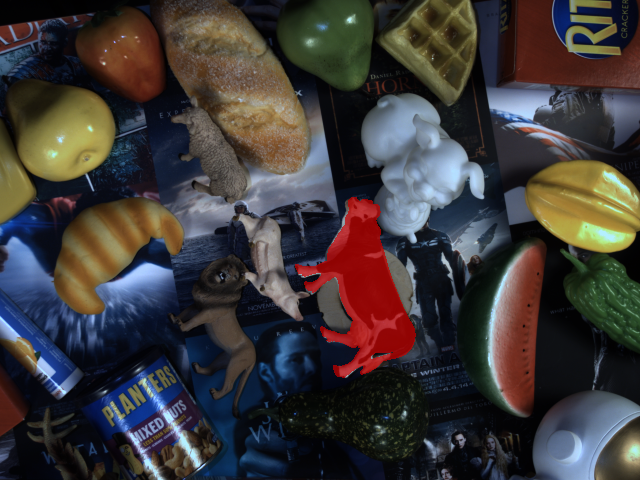}}\end{subfigure}
    \begin{subfigure}{.16\textwidth}{\centering\includegraphics[width=\linewidth]{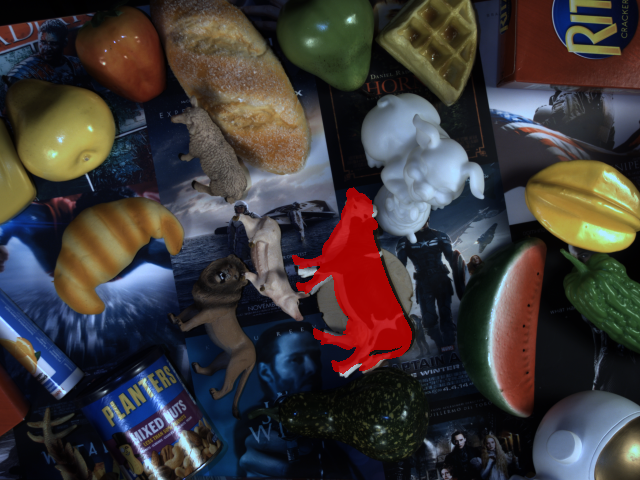}}\end{subfigure}
    
    \begin{subfigure}{.16\textwidth}{\centering\includegraphics[width=\linewidth]{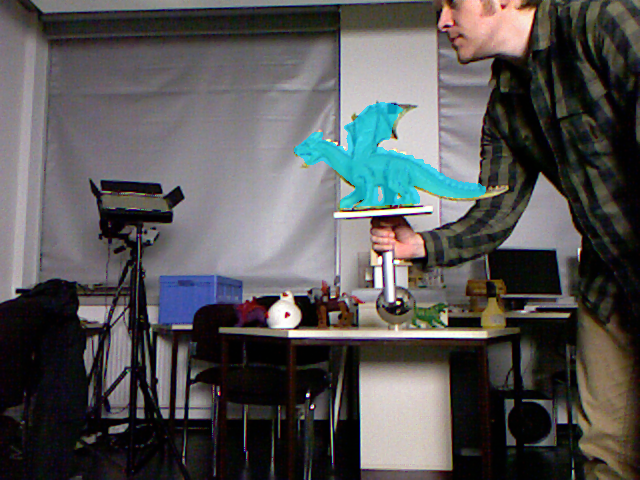}}\end{subfigure}
    \begin{subfigure}{.16\textwidth}{\centering\includegraphics[width=\linewidth]{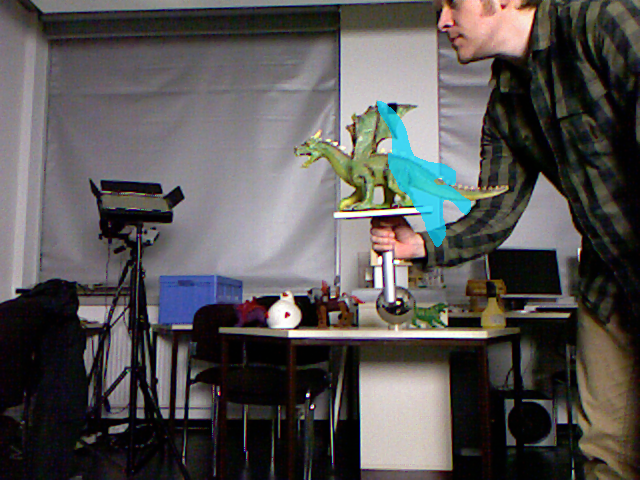}}\end{subfigure}
    \begin{subfigure}{.16\textwidth}{\centering\includegraphics[width=\linewidth]{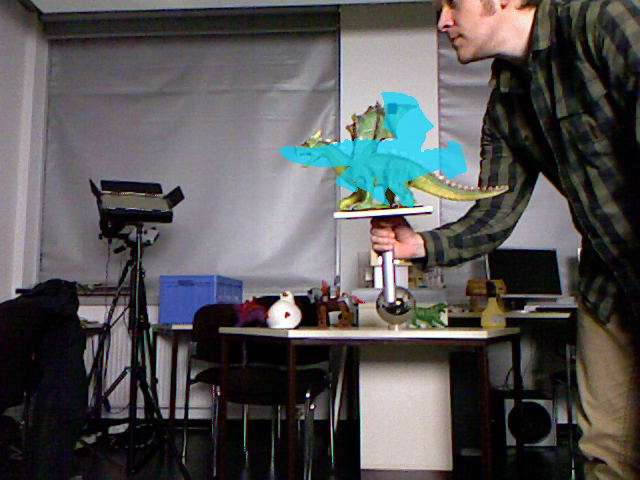}}\end{subfigure}
    \begin{subfigure}{.16\textwidth}{\centering\includegraphics[width=\linewidth]{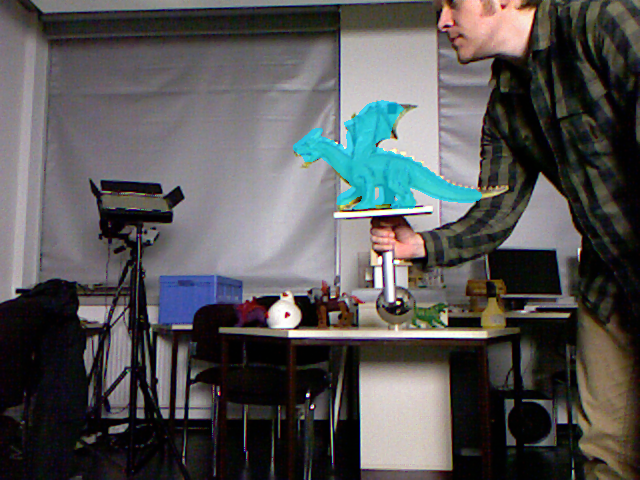}}\end{subfigure}
    \begin{subfigure}{.16\textwidth}{\centering\includegraphics[width=\linewidth]{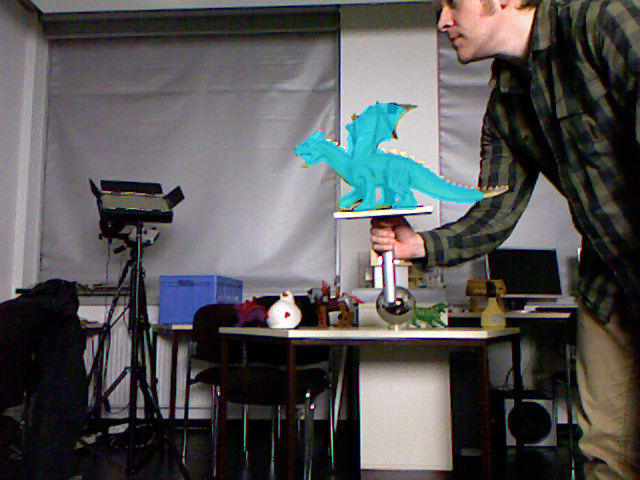}}\end{subfigure}
    \begin{subfigure}{.16\textwidth}{\centering\includegraphics[width=\linewidth]{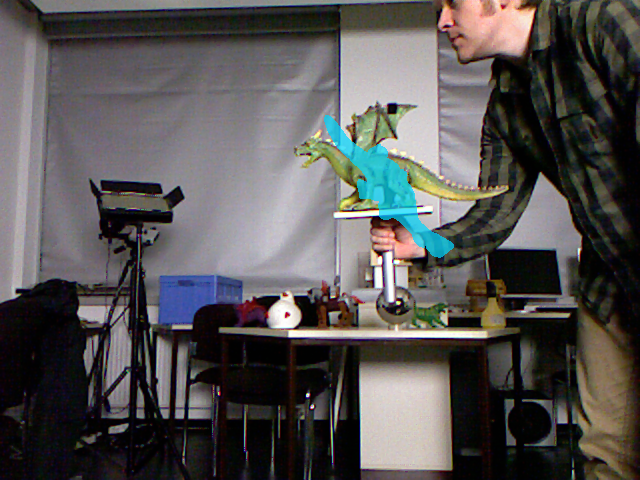}}\end{subfigure}

    \begin{subfigure}{.16\textwidth}{\centering\includegraphics[width=\linewidth]{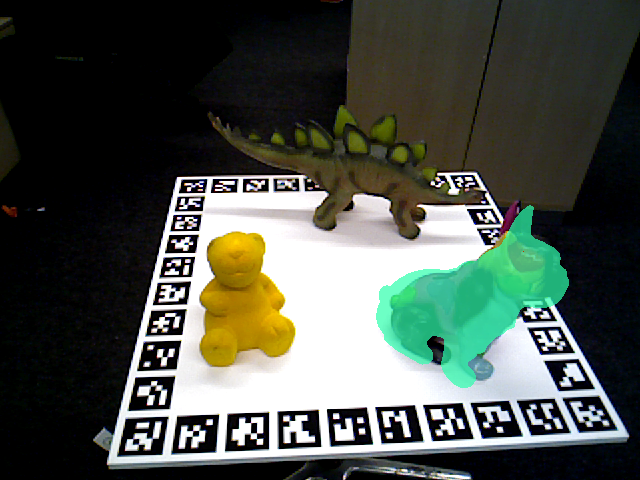}\caption{BPNet(ours)}}\end{subfigure}
    \begin{subfigure}{.16\textwidth}{\centering\includegraphics[width=\linewidth]{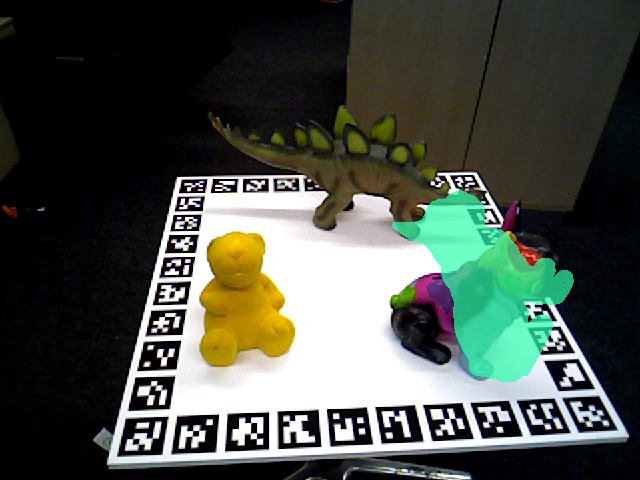}\caption{IDAM(GNN)}}\end{subfigure}
    \begin{subfigure}{.16\textwidth}{\centering\includegraphics[width=\linewidth]{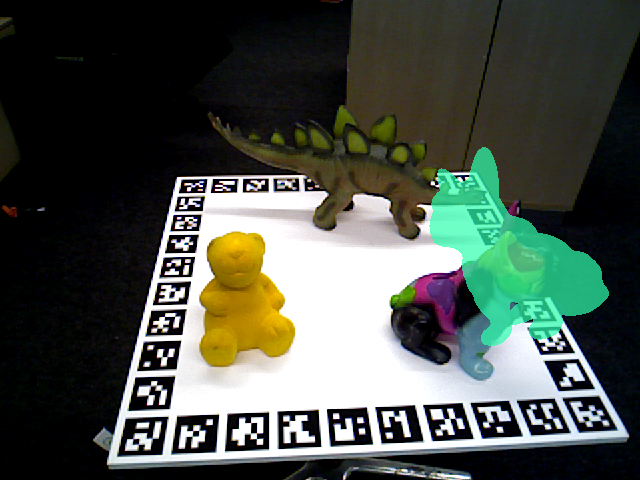}\caption{IDAM(GNN)-M}}\end{subfigure}
    \begin{subfigure}{.16\textwidth}{\centering\includegraphics[width=\linewidth]{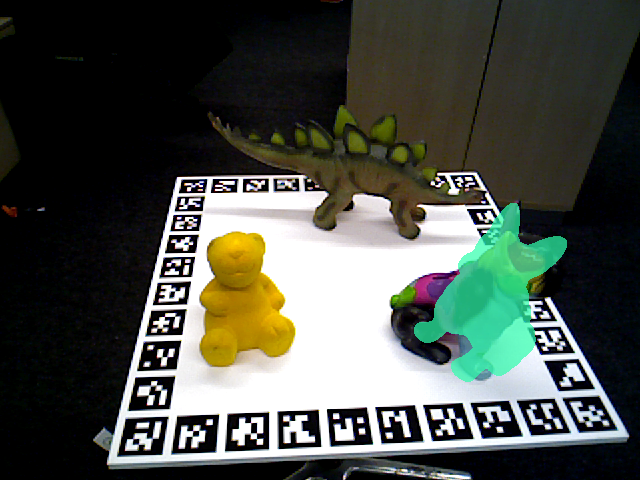}\caption{IDAM(DGCNN-R)}}\end{subfigure}
    \begin{subfigure}{.16\textwidth}{\centering\includegraphics[width=\linewidth]{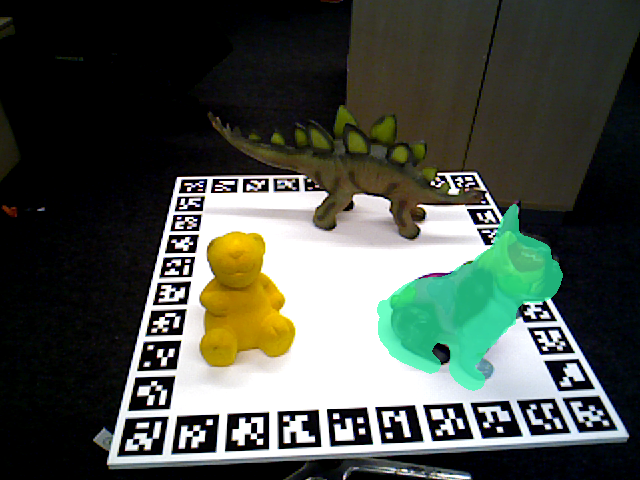}\caption{IDAM(DGCNN-C)}}\end{subfigure}
    \begin{subfigure}{.16\textwidth}{\centering\includegraphics[width=\linewidth]{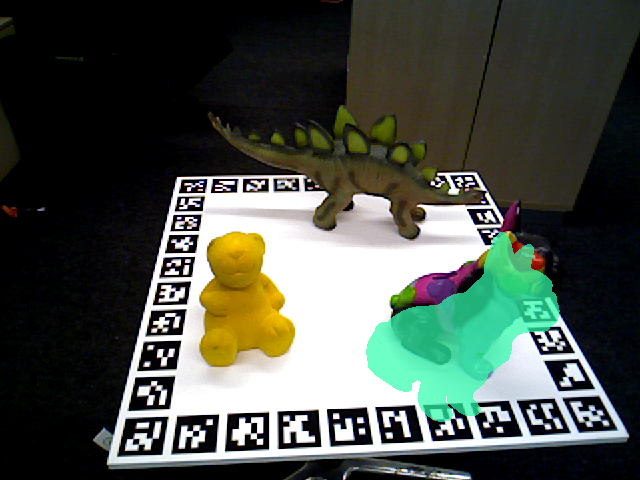}\caption{TEASER++}}\end{subfigure}
    
    \caption{\label{xepose_qual} Qualitative results on the XEPose (top), TUD-L (middle) and HomebrewedDB (bottom) datasets.}
    \vspace{-1.4em}
\end{figure*}

%% file: table/xepose.tex
\begin{table*}[!htb]
    \centering
    {\footnotesize
    \setlength{\tabcolsep}{2.3mm}{
    \begin{tabular}{l|cccccc|cccccc|c}
        \toprule
        & \multicolumn{6}{c|}{Rotation mAP} & \multicolumn{6}{c|}{Translation mAP} & \multicolumn{1}{c}{ADD} \\
        Method & $5^{\circ}$ & $10^{\circ}$ & $15^{\circ}$ & $20^{\circ}$ & $25^{\circ}$ & $30^{\circ}$& $0.5cm$ & $1cm$ & $2cm$ & $5cm$ & $10cm$ & $15cm$ & $0.1d$\\
        \midrule
        ICP         & 0.00 & 0.00 & 0.01 & 0.01 & 0.02 & 0.03 & 0.03 & 0.17 & 0.77 & 1.00 & 1.00 & 1.00 & 0.02 \\
        FGR         & 0.01 & 0.02 & 0.04 & 0.05 & 0.06 & 0.07 & 0.06 & 0.18 & 0.70 & 0.99 & 1.00 & 1.00 & 0.06 \\
        TEASER++    & 0.25 & 0.25 & 0.27 & 0.29 & 0.32 & 0.35 & 0.28 & 0.36 & 0.55 & 0.95 & 1.00 & 1.00 & 0.31 \\ 
        DCP(v2)     & 0.00 & 0.00 & 0.00 & 0.00 & 0.00 & 0.00 & 0.00 & 0.00 & 0.12 & 0.78 & 1.00 & 1.00 & 0.00 \\ 
        PRNet       & 0.00 & 0.00 & 0.00 & 0.01 & 0.02 & 0.03 & 0.00 & 0.02 & 0.12 & 0.73 & 1.00 & 1.00 & 0.00 \\ 
        IDAM(GNN)   & 0.03 & 0.08 & 0.14 & 0.20 & 0.25 & 0.29 & 0.07 & 0.23 & 0.62 & 1.00 & 1.00 & 1.00 & 0.17  \\ 
        \rowcolor{mapillarygreen}
        BPNet       & 0.30 & 0.55 & 0.71 & 0.81 & 0.86 & 0.90 & 0.52 & 0.84 & 1.00 & 1.00 & 1.00 & 1.00 & 0.77\\ 
        \rowcolor{mapillarygreen}
        IDAM(GNN)-M     & 0.06 & 0.15 & 0.23 & 0.30 & 0.35 & 0.41 & 0.16 & 0.36 & 0.72 & 1.00 & 1.00 & 1.00 & 0.29 \\ 
        \rowcolor{mapillarygreen}
        IDAM(DGCNN running BN)-M     & 0.24 & 0.42 & 0.54 & 0.58 & 0.62 & 0.64 & 0.42 & 0.66 & 0.84 & 1.00 & 1.00 & 1.00 & 0.60 \\ 
        \rowcolor{mapillarygreen}
        IDAM(DGCNN current BN)-M     & 0.33 & 0.54 & 0.62 & 0.65 & 0.66 & 0.68 & 0.49 & 0.69 & 0.87 & 1.00 & 1.00 & 1.00 & 0.63 \\ 
        \bottomrule
    \end{tabular}
    }}
    \caption{Quantitative comparison of our method and the baselines on the \textbf{XEPose} real scene dataset.}
    \label{tab:xepose}
    \vspace{-1.25em}
\end{table*}

%% file: tex/4-experiment.tex
\section{Experiments}
\label{sec:experiments}
\input{table/tudl}
\input{table/hb}
To verify the effectiveness of our guidelines, we apply them to another baseline method, IDAM~\cite{Li20}. Furthermore, we demonstrate the generality of our analysis by performing additional experiments on the real TUD-L and HomeBrew datasets, captured with consumer-grade depth sensors.

\subsection{Making IDAM Work on Real Data}
\label{sec:xepose}
\paragraph{Implementation Details and Baselines.}
IDAM was developed after DCP and is therefore already more robust. In particular, IDAM employs an outlier rejection strategy similar to the one used in PRNet, and replaced the SVD+MSE loss with an NLL-based loss function, thus stabilizing the training process. These improvements are all in line with the guidelines that we proposed in \Cref{sec:outlier_rejection,sec:nll_loss}.
Nevertheless, we can still improve IDAM in the following ways. First, note that IDAM is a keypoint-based method that takes as input a feature embedding, and was demonstrated in~\cite{Li20} with two different embeddings, GNN and FPFH. In both cases, we can apply our data processing and voxel downsampling strategies. Furthermore, we replaced the GNN with a DGCNN embedding, and tested both running and current statistics in its batch-normalization layers. Note that, in both cases, we used our data processing and voxel downsampling techniques.
We refer to the resulting networks as IDAM-M, further indicating the embedding. We defer the FPFH embedding to the supplementary material, as its accuracy was lower that the others.

In addition to comparing IDAM to IDAM-M, we also report the results of traditional methods, ICP~\cite{Besl92} and FGR~\cite{Zhou16}, for which we use the Open3D implementation, as well as TEASER++, using the publicly available code. Furthermore, we provide the results of the state-of-the-art learning-based methods, DCP~\cite{Wang19e} and PRNet~\cite{Wang19f}, obtained using their respective github code. Finally, we also compare these results to those of our BPNet.

\vspace{-1.25em}
\paragraph{Training Data.}
To beneﬁt from having access to virtually inﬁnite amounts of training data, we use Pytorch3D to generate on-the-fly scenes that contain individual objects without background. To this end, we use the mesh models from the ModelNet40 training split and the given camera intrinsic parameters and camera pose to generate depth maps. 
We then convert these depth maps to target point clouds. The source point clouds are obtained by uniformly sampling the mesh models' surface. 

To build the ground-truth assignment matrix $\bM$ used in the NLL loss function,
we transform the source point set $\bX$ using the ground-truth transformation $\bT$, giving us $\btX$. We then compute the pairwise Euclidean distance matrix between $\btX$ and the target point set $\bY$, which we threshold to $0.05$ to obtain a correspondence matrix $\bM \in \{0, 1\}$. We augment $\bM$ with an extra row and column acting as outlier bins to obtain $\bbM$. The points without any correspondence are treated as outliers, and the corresponding positions in $\bbM$ are set to one. This  does not guarantee a bijective matching, which we address via a forward-backward check. 

In the evaluation process, we use the model trained on the ModelNet40 training split without any fine-tuning for all the methods, as in  Section~\ref{eval_protocols}. 

\vspace{-1.25em}
\paragraph{Results.}
As shown in Table~\ref{tab:xepose} DCP and PRNet fail completely on the XEPose dataset. By contrast, because the vanilla IDAM(GNN) uses a feature reduction operation in the GNN whose principle is consistent with our data processing strategies, it reaches a $17\%$ ADD-0.1. After applying our guidelines to IDAM(GNN), the result is improved to $29\%$ in ADD-0.1. IDAM(DGCNN)-M yields even better results, particularly when using the current BN statistics.

Compared to the keypoint selection strategy used in IDAM, the Sinkhorn (hard selection) approach exploited in our BPNet better copes with the noise in the real data. As a result, our BPNet produces the best performance of $77\%$ in ADD-0.1. 

In Figure~\ref{xepose_qual}, we provide qualitative results for the methods that yield reasonable results, projecting the model in the predicted pose in images for visualization purpose. The poses obtained by our BPNet match much more closely the ground-truth ones than those produced by the baselines.

\subsection{Generalization to Other Real Datasets}
\label{tudl_hb}
\paragraph{TUD-L.}
The TUD-L dataset was captured using a consumer-grade depth sensor of $2mm$ precision. It contains training and testing image sequences that show three moving objects under eight lighting conditions. The TUD-L dataset is more challenging than the XEPose dataset because of its reduced sensor accuracy and higher noise. This also means a larger gap w.r.t. the synthetic training data.

As a consequence, and as shown in Table~\ref{tab:tudl}, 
most methods fail on this dataset. Among the baselines, only TEASER++ and IDAM show non-zero ADD-0.1 on the dataset, which are far from working well. After performing our guidelines, the IDAM(DGCNN current BN)-M gives an accuracy up to $36\%$, seven times that of the vanilla IDAM. 
As for our BPnet, it yields an accuracy of $67\%$ in ADD-0.1, nearly twice that of the second best method.
\vspace{-1.25em}
\paragraph{HomebrewedDB.}
The HomebrewedDB dataset contains 13 scenes captured with the same depth sensor as TUD-L. Besides 25 household objects, HomebrewedDB contains 8 industrial objects. This dataset comprises challenges such as occlusions, changing light conditions and changes in object appearance. Furthermore, in contrast to the other two datasets, it contains symmetric objects that are more difficult to handle because they create ambiguities.

As shown in Table~\ref{tab:hb}, most existings methods suffer from the challenges in this dataset and thus fail to give satisfactory results, only TEASER++ gives $7\%$ in ADD-0.1. Our BPNet and IDAM(DGCNN current BN)-M give the best results, with an ADD-0.1 accuracy of $13\%$ and $14\%$. The third best method is IDAM(DGCNN running BN)-M, which is improved by the data normalization and voxel downsampling, reaching an accuracy of $9\%$.

In the two bottom rows of Figure~\ref{xepose_qual}, we show qualitative results for TUD-L and HomebrewedDB. As in the XEPose case, the results obtain with our BPNet better match the ground-truth poses.
\vspace{-.5em}

%% file: table/tudl.tex
\begin{table*}[!htb]
    \centering
    {\footnotesize
    \setlength{\tabcolsep}{2.2mm}{
    \begin{tabular}{l|cccccc|cccccc|c}
        \toprule
        & \multicolumn{6}{c|}{Rotation mAP} & \multicolumn{6}{c|}{Translation mAP} & \multicolumn{1}{c}{ADD} \\
        Method & $5^{\circ}$ & $10^{\circ}$ & $15^{\circ}$ & $20^{\circ}$ & $25^{\circ}$ & $30^{\circ}$& $0.5cm$ & $1cm$ & $2cm$ & $5cm$ & $10cm$ & $15cm$ & $0.1d$\\
        \midrule
        ICP         & 0.00 & 0.00 & 0.00 & 0.00 & 0.00 & 0.00 & 0.00 & 0.04 & 0.15 & 0.56 & 0.99 & 1.00 & 0.00 \\
        FGR         & 0.00 & 0.00 & 0.00 & 0.00 & 0.00 & 0.00 & 0.00 & 0.03 & 0.13 & 0.59 & 0.98 & 1.00 & 0.00 \\
        TEASER++    & 0.02 & 0.03 & 0.04 & 0.04 & 0.05 & 0.06 & 0.02 & 0.03 & 0.06 & 0.25 & 0.70 & 0.94 & 0.04 \\ 
        DCP(v2)     & 0.00 & 0.00 & 0.00 & 0.00 & 0.00 & 0.00 & 0.00 & 0.00 & 0.03 & 0.13 & 0.43 & 0.60 & 0.00 \\ 
        PRNet     & 0.00 & 0.00 & 0.00 & 0.01 & 0.01 & 0.02 & 0.00 & 0.00 & 0.01 & 0.24 & 0.81 & 0.99 & 0.00 \\ 
        IDAM(GNN)   & 0.01 & 0.03 & 0.05 & 0.08 & 0.09 & 0.11 & 0.00 & 0.01 & 0.07 & 0.45 & 0.88 & 0.99 & 0.05  \\ 
        \rowcolor{mapillarygreen}
        BPNet & 0.29 & 0.51 & 0.66 & 0.74 & 0.77 & 0.79 & 0.00 & 0.08 & 0.43 & 0.94 & 1.00 & 1.00 & 0.67 \\
        \rowcolor{mapillarygreen}
        IDAM(GNN)-M     & 0.02 & 0.06 & 0.09 & 0.12 & 0.14 & 0.16 & 0.00 & 0.02 & 0.10 & 0.51 & 0.88 & 0.99 & 0.09 \\ 
        \rowcolor{mapillarygreen}
        IDAM(DGCNN running BN)-M     & 0.11 & 0.21 & 0.31 & 0.40 & 0.45 & 0.48 & 0.00 & 0.04 & 0.19 & 0.72 & 0.98 & 1.00 & 0.31 \\ 
        \rowcolor{mapillarygreen}
        IDAM(DGCNN current BN)-M    & 0.19 & 0.27 & 0.32 & 0.35 & 0.38 & 0.41 & 0.04 & 0.16 & 0.38 & 0.67 & 0.99 & 1.00 & 0.36 \\ 
        \bottomrule
    \end{tabular}
    }}
    \caption{Quantitative comparison of our method and the baselines on the \textbf{TUD-L} real scene dataset.}
    \label{tab:tudl}
    \vspace{-.5em}
\end{table*}

%% file: table/hb.tex
\begin{table*}[!htb]
    \centering
    {\footnotesize
    \setlength{\tabcolsep}{2.2mm}{
    \begin{tabular}{l|cccccc|cccccc|c}
    
        \toprule
        & \multicolumn{6}{c|}{Rotation mAP} & \multicolumn{6}{c|}{Translation mAP} \\
        Method & $5^{\circ}$ & $10^{\circ}$ & $15^{\circ}$ & $20^{\circ}$ & $25^{\circ}$ & $30^{\circ}$& $0.5cm$ & $1cm$ & $2cm$ & $5cm$ & $10cm$ & $15cm$ & $0.1d$\\
        \midrule
        ICP         & 0.00 & 0.00 & 0.00 & 0.00 & 0.00 & 0.00 & 0.00 & 0.01 & 0.19 & 0.94 & 1.00 & 1.00 & 0.00 \\
        FGR         & 0.00 & 0.00 & 0.00 & 0.01 & 0.01 & 0.02 & 0.00 & 0.02 & 0.19 & 0.90 & 1.00 & 1.00 & 0.01 \\
        TEASER++    & 0.03 & 0.05 & 0.06 & 0.07 & 0.09 & 0.11 & 0.07 & 0.11 & 0.21 & 0.61 & 0.94 & 0.99 & 0.07 \\ 
        DCP(v2)     & 0.00 & 0.00 & 0.00 & 0.00 & 0.01 & 0.01 & 0.00 & 0.00 & 0.01 & 0.38 & 0.90 & 0.98 & 0.00 \\ 
        PRNet      & 0.00 & 0.00 & 0.00 & 0.01 & 0.01 & 0.02 & 0.00 & 0.00 & 0.02 & 0.36 & 0.91 & 0.99 & 0.00 \\  
        IDAM(GNN)   & 0.00 & 0.01 & 0.02 & 0.03 & 0.04 & 0.05 & 0.00 & 0.02 & 0.14 & 0.69 & 0.98 & 1.00 & 0.01 \\ 
        \rowcolor{mapillarygreen}
        BPNet     & 0.03 & 0.12 & 0.23 & 0.32 & 0.39 & 0.44 & 0.00 & 0.03 & 0.28 & 0.92 & 0.99 & 1.00 & 0.13 \\
        \rowcolor{mapillarygreen}
        IDAM(GNN)-M     & 0.00 & 0.01 & 0.03 & 0.04 & 0.05 & 0.07 & 0.01 & 0.04 & 0.17 & 0.70 & 0.98 & 1.00 & 0.02 \\ 
        \rowcolor{mapillarygreen}
        IDAM(DGCNN running BN)-M     & 0.02 & 0.06 & 0.09 & 0.13 & 0.16 & 0.18 & 0.01 & 0.07 & 0.25 & 0.77 & 0.98 & 1.00 & 0.09 \\ 
        \rowcolor{mapillarygreen}
        IDAM(DGCNN current BN)-M     & 0.02 & 0.07 & 0.12 & 0.16 & 0.19 & 0.22 & 0.02 & 0.11 & 0.36 & 0.85 & 0.99 & 1.00 & 0.14 \\ 
        \bottomrule
    \end{tabular}
    }}
    \caption{Quantitative comparison of our method and the baselines on the \textbf{HomebrewedDB} real scene dataset.}
    \label{tab:hb}
    \vspace{-1.25em}
\end{table*}

%% file: tex/5-conclusion.tex
\section{Conclusion}
We have identified the factors that prevent existing learning-based 3D registration methods from working on real-world data and proposed solutions to address these limitations. We have summarized our findings into guidelines and demonstrated their effectiveness on different baseline methods (i.e., DCP, IDAM) and different datasets (i.e., XEPose, TUD-L and HomebrewedDB).
As a result, our best practice network, BPNet, has demonstrated a remarkable ability to generalize to real data and previously-unseen objects.
As such, it provides a strong baseline to benchmark future 3D registration methods on real-world data. We also expect our guidelines to help others to improve the convergence speed and robustness of their networks for deployment with real data; as evidenced by our results on the HomebrewedDB dataset, this problem is not yet solved.